\pdfoutput=1

\documentclass[11pt]{article}

\usepackage[final]{coling}

\usepackage{times}
\usepackage{latexsym}

\usepackage[T1]{fontenc}

\usepackage[utf8]{inputenc}

\usepackage{microtype}

\usepackage{inconsolata}

\usepackage{graphicx}

\usepackage{booktabs}
\usepackage{multirow}
\usepackage{amsfonts}
\usepackage{amsmath}
\usepackage{colortbl}
\usepackage{array}
\usepackage{caption}
\usepackage{listings}
\usepackage{xspace}
\usepackage{ifthen}

\newcommand{\xslue}{x\texttt{SLUE}\xspace}
\newcommand{\lisa}{\textsc{Lisa}\xspace}
\newcommand{\llama}{\texttt{llama3-8b}\xspace}
\newcommand{\gptthree}{\texttt{gpt-3.5-turbo}\xspace}

\title{Latent Space Interpretation for Stylistic Analysis and Explainable Authorship Attribution}

\author{
 \textbf{Milad Alshomary\textsuperscript{\dag}},
 \textbf{Narutatsu Ri\textsuperscript{\dag}},
 \textbf{Marianna Apidianaki\textsuperscript{\ddag}},
 \\
 \textbf{Ajay Patel\textsuperscript{\ddag}},
 \textbf{Smaranda Muresan\textsuperscript{\dag}},
 \textbf{Kathleen McKeown\textsuperscript{\dag}}
\\
\\
 \textsuperscript{\dag}Columbia University, New York, NY
 \\
 \textsuperscript{\ddag}University of Pennsylvania, Philadelphia, PA
\\
 \small{
   \textbf{Correspondence:} \href{ma4608@columbia.edu}{ma4608@columbia.edu}
 }
}

\RequirePackage{type1cm}
\RequirePackage{color}
\RequirePackage{soul}
\setstcolor{blue}
\definecolor{violet}{rgb}{0.5,0.0,0.5}
\newsavebox\bscombox
\newcommand{\bscom}[3][]{%
	\sbox{\bscombox}{\fontsize{8}{9}\selectfont#1#2#3}
	\noindent
	\st{#2}{\selectfont
		\color{blue}#3\ifx\\#1\\\else{\fontsize{8}{9}\selectfont\color{violet}[#1]}\fi
	}
}

\newsavebox\scrcombox
\newcommand{\scrcom}[3][]{%
    \sbox{\scrcombox}{\fontsize{8}{9}\selectfont#1#2#3}
    \noindent
    {\selectfont\color{lightgray}#2}
    \ifx\\#1\\\else{\fontsize{8}{9}\selectfont\color{orange}[\textbf{ER}: #1]}\fi
    {\selectfont\color{blue}#3} 
}

\begin{document}
\maketitle

\begin{abstract}
Recent state-of-the-art authorship attribution methods learn authorship representations of texts in a latent, non-interpretable space, hindering their usability in real-world applications. Our work proposes a novel approach to interpreting these learned embeddings by identifying representative points in the latent space and utilizing LLMs to generate informative natural language descriptions of the writing style of each point. We evaluate the alignment of our interpretable space with the latent one and find that it achieves the best prediction agreement compared to other baselines. Additionally, we conduct a human evaluation to assess the quality of these style descriptions, validating their utility as explanations for the latent space. Finally, we investigate whether human performance on the challenging AA task improves when aided by our system's explanations, finding an average improvement of around +20\% in accuracy.
\end{abstract}

\section{Introduction}

The task of authorship attribution (AA) involves identifying the author of a document by extracting stylistic features and comparing them with those found in documents by other authors. 
AA is of particular interest due to its real-world applications, such as forensic linguists relying on these systems when testifying in criminal and civil trials \cite{tiersma:2024}. 
Given the sensitivity of such tasks, ensuring that a model’s predictions can be verified through clear, distilled explanations is crucial for building user trust \cite{toreini:2020}.

Early approaches to authorship attribution focused on identifying writing styles as features and training classifiers on these features to capture document-author similarities \cite{koppel:2004}. 
While inherently interpretable, these methods fall short in performance compared to recent state-of-the-art approaches based on transformer architectures, which match authors to documents based on vector similarities in a learned latent space \cite{rivera-soto-etal-2021-learning, wegmann-etal-2022-author}. 
These deep learning-based methods, however, inherently operate in a black-box fashion, and explaining their predictions is challenging. While a rich body of literature exists on interpreting deep learning models, research specifically exploring the explainability of AA models is nascent, and existing studies on style representations \cite{wegmann-etal-2022-author, lyu:2023, erk:2024} do not address their application for explainability---a research gap we aim to fill.

\begin{figure}
    \centering
    \includegraphics{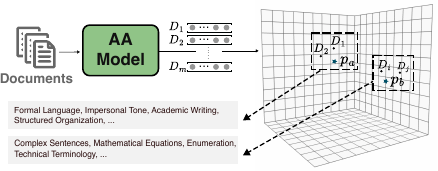}
    \caption{Our approach for explaining authorship attribution predictions. We identify $k$ clusters with centroids $p_1, \dots, p_k$ in the embedding space and associate each with writing style features. The writing style of a document $D_i$ is explained by aggregating the style features of its closest cluster.}
    \label{fig:main_approach_idea}
\end{figure}


In this paper, we hypothesize that embedding-based authorship attribution models learn to map texts into regions in the latent space that represent specific writing styles. 
By locating a document with respect to these regions in the latent space, we can identify its style features relevant to the model's representation, which can serve as useful explanations to assist humans in the authorship attribution task. 
Our proposed approach identifies these relevant regions and their corresponding writing style by clustering a set of training documents in the latent space and assigns each cluster a distribution over style features generated by prompting large language models (LLMs). 
The centroids of these clusters then serve as the basis for a new interpretable space. 
Given a new document $D_i$, we first identify its top $k$ similar clusters, and we then aggregate a set of style features from these clusters' style representation to describe its writing style. This is illustrated in the toy example in Figure~\ref{fig:main_approach_idea}, where two clusters are identified in the latent space ($p_a$ and $p_b$), each with its distinct list of style features. The style of a document $D_1$ can be explained by the set of features of its most similar cluster $p_a$.

We conduct automatic and manual evaluations to test our hypothesis that the clusters' features can describe the writing style of unseen documents and are useful to the authorship attribution task.  
In our automatic evaluation, we measure the agreement of predictions made in our constructed interpretable space and in the original model's latent space, and we compare that against other baselines. 
Our approach achieves the highest Pearson correlation of 0.79, surpassing baseline methods that range from 0.2 to 0.4. We also conduct a human evaluation to verify whether the style features associated with the identified clusters reflect the writing style of unseen documents. 
The results indicate that this is indeed the case since humans rank these style features higher than other non-associated ones in 72\% of the cases. 
Finally, we measure the usefulness of our explanations for the authorship attribution task by asking participants to identify the author of new documents with and without having access to our explanations.
Our explanations improved their agreement and increased the annotators' accuracy in the task by an average of 20\%.


In summary, our contributions are as follows:
\begin{itemize}
    \itemsep 0em
    \item A novel approach to interpreting the latent space of embedding-based AA models.
    \item Experimental evidence demonstrating the validity of style descriptions generated by our interpretable space.
    \item Demonstration of the utility of the interpretable space's explanations for the AA task.
\end{itemize}
\section{Related Work}
\paragraph{Authorship Attribution}
Early approaches to authorship attribution focused on modeling linguistic features such as syntactic structure and function word frequencies to capture similarities in writing style \citep{koppel:2004}. 
Recently, transformer-based models have achieved state-of-the-art results by fine-tuning on large corpora to learn embeddings that reflect different writing styles. 
For instance, \citet{rivera-soto-etal-2021-learning} proposed a contrastive objective function that maps documents written by the same author into vectors situated closer in the embedding space. 
\citet{wegmann-etal-2022-author} introduced an approach to ensure that the embedding space of an authorship attribution model represents style rather than content. 
\citet{tyo:2022} provide a comprehensive survey of approaches for the authorship attribution task.

\paragraph{Interpreting Embedding Spaces}
Despite their strong performance, transformer-based models learn latent representations that are not interpretable, making the models less explainable. 
A well-established line of research focuses on methods to interpret the learned embeddings. For example, \citet{simhi:2023} proposed an approach that maps Wikipedia concepts into the latent space and uses these mapped embeddings as dimensions of an interpretable space. Few works have studied the interpretability of style embeddings learned for the authorship attribution task. 
\citet{wegmann-etal-2022-author} investigated the presence of specific style features in the representations learned (or induced) by their model. 
\citet{lyu:2023} showed that it is possible to identify vectors representing lexical stylistic features (such as complexity and formality) in the latent space of pre-trained language models. 
However, these studies do not extend beyond small-scale analyses to probe embeddings for specific style features. The usefulness of this knowledge for explainable authorship attribution is unexplored.

\paragraph{Explainability}
General frameworks such as LIME \cite{ribeiro:2016} and SHAP \cite{lundberg:2017} aim to explain
the behavior of deep learning models.
These frameworks measure the contribution of single features (tokens) to the final prediction. Inferring human-level explanations from these scores might be easy for simple classification tasks such as sentiment analysis. This becomes more complicated when dealing with complex tasks such as authorship attribution, where predictions might rely on features beyond the level of tokens (e.g., syntax or discourse-related features).
\section{Explaining Authorship Attribution}
\label{sec:approach}


\subsection{Problem Setup}
In this study, we frame the authorship attribution task as follows: Given a collection of $n$ documents ${D = \{D_1, \dots, D_n\}}$ written by $m$ different authors $A_1, \dots, A_m$, predict how likely it is for two documents to be written by the same author. Authorship attribution methods typically learn a function $f(\cdot)$ that maps documents into a latent embedding space, and then they rely on it to predict the author of new documents at inference time \citep{rivera-soto-etal-2021-learning, wegmann-etal-2022-author}. 
But these models are opaque and not interpretable. 

A natural explanation of the underlying mechanism of $f(\cdot)$ is that the models learn to associate specific style features with regions in the latent space. An intuitive approach to uncovering this underlying mechanism --- in line with the approach proposed by \citet{simhi:2023} for interpreting general learned embeddings --- would depart from texts with specific predefined style features and aim at identifying their corresponding latent representations.
This approach is limited by the need for predefined features and the strong assumption that the AA model represents them. 
In contrast to this ``top-down'' method, our approach to interpreting the latent space can be described as ``bottom-up.'' As detailed below, instead of using a predefined set of features, we automatically locate salient regions in the latent space that are relevant to the model's predictions and map them to an automatically discovered set of style features. This ensures that the latent regions and the style features are both relevant to the model's prediction.

\begin{figure}
    \centering
    \includegraphics{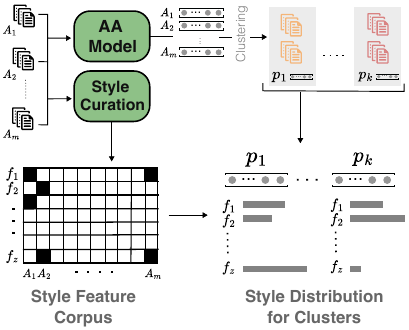}
    \caption{Our approach: Given the training corpus $D^\textup{train}$ with documents from authors $A_1, \dots, A_m$, we generate style descriptions for each document to construct the style corpus. We then identify relevant regions in the latent space $p_1, \dots, p_k$ by clustering author-level representations and aggregate style features to obtain style representations for each region.}
    \label{fig:constructing_interp_space}
\end{figure}

\subsection{Latent Space Interpretation}\label{sec:latent_space_interpretation}
We propose a two-step procedure for interpreting the latent space of AA models. Our approach is illustrated in Figure~\ref{fig:constructing_interp_space} and explained below. 

\paragraph{Identifying Representative Points}
Given a training set $D^\textup{train}$, containing author-labeled texts, we first identify $k$ representative points ($p_1, \dots, p_k$) in the latent space that are relevant to the AA model's predictions.
For this, we first obtain author-level representations (which we call hereafter author embeddings) by averaging the representations of documents by the same author. 
Although each author embedding could be considered as a single representative point, one can obtain better style representations by grouping similar authors since more documents will be available in each cluster. Therefore, we cluster the author embeddings and take their centroids as our final $k$ representative points. 

\paragraph{Mapping to Style Distributions} 
We consider the $k$ representative points identified in the latent space as our interpretable space dimensions. 
This step aims to map each point and its associated documents to the corresponding style features. To this end, we automatically construct a set of style features describing all the training documents. Then, we map each point to a distribution over these style features given its associated documents.

\begin{figure}[t]
	\centering
	\includegraphics{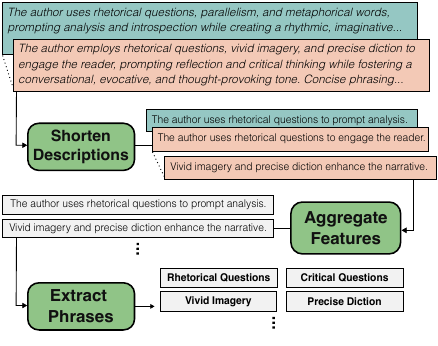}
	\caption{Refinement steps applied on style description distilled from \llama.}
	\label{fig:style_features_transformation} 
\end{figure}

We build this set of features following \citet{patel-etal-2023-learning}, who used LLMs to describe the writing style of a document in a zero-shot setting. Concretely, we first generate style descriptions for each document in the training set by prompting the LLM. However, the generated descriptions are quite lengthy, as shown in the examples in Figure~\ref{fig:style_features_transformation}, and they present substantial overlaps. Therefore, as exemplified in the figure, we process the generated style descriptions as follows: we first prompt LLMs to shorten each of the descriptions. Second, we perform automatic feature aggregation to merge descriptions with similar styles. For this, we construct a pairwise similarity matrix by computing the \emph{Mutual Implication Score} \citep{babakov-etal-2022-large} between each pair of descriptions and merge the ones that are sufficiently similar. Finally, we shorten the aggregated descriptions further by extracting key phrases present therein. These constitute the final set of style features used in our experiments. Implementation details about the LLMs and prompts used are provided in Section \ref{sec:experiment_setup}.

Using the constructed set of style features, we generate a distribution over the generated style features for each representative point ($p_1, \dots, p_k$). To do this, we compute the frequency of each style feature (e.g., vivid imagery, rhetorical questions, etc) in the documents that are associated with the representative points, normalized by the frequency of the feature in the entire training set.

\paragraph{Explaining Model Predictions}
\label{sec:explainauthorship attributionpredictions}
The constructed interpretable space, where each basis is associated with a style distribution, can be used for explainable AA in two ways. First, to explain how the AA model encodes the writing style of a single document, we first project its latent embedding into the interpretable space by computing its cosine similarity to each of the representative points in the latent space. We then explain its writing style by relying on its top $N$ (i.e. the closest) representative points. Second, to explain why the AA model predicts that two documents were written by the same author, we similarly project both documents into the interpretable space and generate a unified style description by aggregating the top representative points from both documents.
\section{Experiment Setup}\label{sec:experiment_setup}

In this section, we present the authorship attribution dataset and the models used in our experiments. We also explain the implementation details of our approach and the baselines used for comparison.

\begin{table}[t]
    \centering
    \setlength{\tabcolsep}{2pt}
    \scalebox{0.9}{
    \begin{tabular}{lrrr}
        \toprule
        \textbf{Statistic} & \textbf{Train} & \textbf{Dev} & \textbf{Test} \\
        \midrule
        \# Documents & 15822 & 2456 &  6107\\
        \# Authors & 4142 & 635 & 1586 \\
        Avg. documents per Author & 3.8 & 3.8 & 3.8 \\
        \bottomrule
    \end{tabular}}
    \caption{Statistics for the training, development, and test splits of the HRS dataset.}
    \label{tab:dataset_stats}
\end{table}

\begin{figure}
    \centering
    \includegraphics[width=0.48\textwidth]{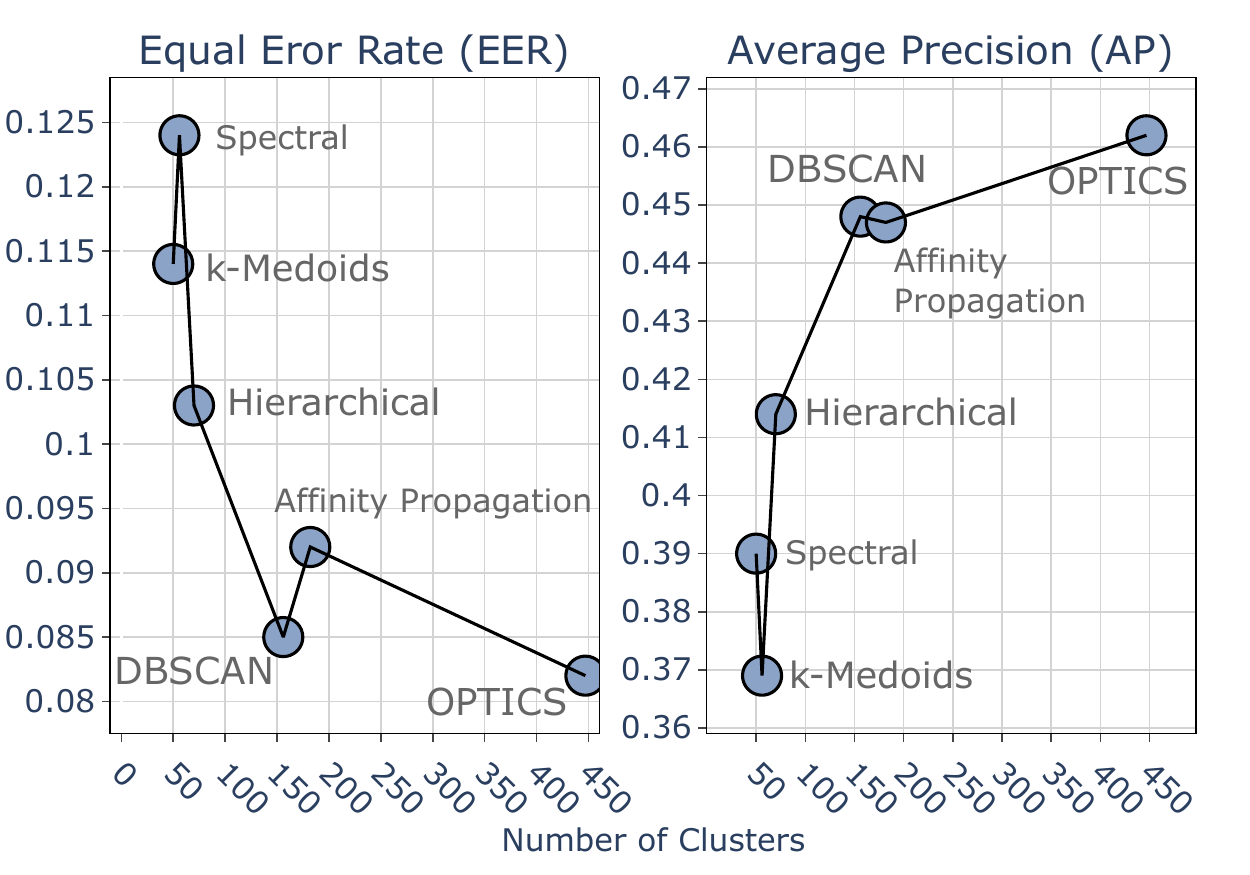}
    \caption{\label{fig:clustering_performance} Performance comparison of cluster assignments by number of clusters. Smaller EER and larger AP indicate better performance. Note that both metrics naturally favor assignments with more clusters.}
\end{figure}

\subsection{Dataset \& AA Models} 
\paragraph{Dataset} To evaluate our approach, we utilize an authorship corpus compiled for the purposes of the IARPA HIATUS research program.\footnote{The corpus was compiled by the Test \& Evaluation (T\&E) team of the HIATUS program and distributed to the participants for training and testing their models.} The corpus comprises documents from 5 online discussion forums spanning different domains: BoardGameGeek, Global Voices, Instructables, Literature StackExchange, and all StackExchanges for STEM topics (physics, computer science, mathematics, etc.). Each source contains documents (posts) tagged with authorship information in their metadata. Sensitive author metadata (PII information) has been removed from the corpus using the Presido analyzer \citep{MsPresidio}, which finds and replaces sensitive entities with placeholders (e.g.,  ``\texttt{PERSON}''). The authorship information of these documents is preserved by assigning random UUIDs to each author. We refer to this dataset as the HRS dataset, and we will share it publicly upon request.\footnote{\url{https://www.iarpa.gov/research-programs/hiatus}} We use the ``cross\_genre'' portion of the HRS dataset, which contains documents written by the same author across different genres (online discussion forums). The cross-genre setting makes authorship attribution more challenging \citep{rivera-soto-etal-2021-learning}. Statistics for the number of documents and authors in each of the training, validation, and test splits are given in Table \ref{tab:dataset_stats}.%


\begin{table}[t!]
    \setlength{\tabcolsep}{8pt}    
    \centering
    \begin{tabular}{lrr}
        \toprule
        \bf AA Model & \bf EER ($\downarrow$) & \bf AP ($\uparrow$) \\ 
        \midrule
        \bf LUAR & 0.06 & 0.49 \\
        \bf FT-LUAR & \bf 0.02 & \bf 0.59 \\
        \bottomrule
    \end{tabular}
    \caption{\label{tab:aa_models_performance} Performance of the AA 
    models on the HRS dataset.}
\end{table}

\paragraph{AA Models} 
We use the state-of-the-art authorship attribution model LUAR \cite{rivera-soto-etal-2021-learning} and a variant that we built using the SBERT library \cite{reimers:2019}. The variant has LUAR as the backbone, which serves to generate the initial text embedding, and a dense layer of 128 dimensions on top to create the final embedding space. We fine-tuned this model on the training split of the HRS dataset using pairs of documents by the same author as positive pairs and the MultipleNegativesRanking loss as the objective function. We train the model for one epoch with a batch size of 48. All other training parameters are left to their default value per the library. At test time, the likelihood of two documents being written by the same author is computed by their cosine similarity in the learned space.

We evaluate the performance of the two authorship attribution models on the test split of the HRS dataset in terms of Equal Error Rate (EER) \citep{meng2019adversarial} and Average Precision (AP). Table \ref{tab:aa_models_performance} presents the results for the two models. Not surprisingly, the FT-LUAR benefits from fine-tuning on the HRS training split, achieving a better EER than LUAR (0.02 vs. 0.06).  These two models are the black boxes whose predictions we would like to explain using our approach.

\subsection{Implementation of Our Approach}
As described, our approach involves (1) identifying representative points by clustering similar authors and (2) automatically constructing a style feature set to assign each of these representative points a distribution over these features. The following describes the implementation details of these steps.

\paragraph{Clustering Similar Authors}
This step serves to identify representative points in the latent space. To find the best clustering, we experiment with various algorithms and parameter values. 
We run each evaluated algorithm 
on the author embeddings from the training set, and use the centroids of the resulting clusters 
as dimensions of corresponding interpretable space. 
Then, to predict how likely it is for a pair of documents in the development set to be written by the same author, we project their latent representations into this evaluated interpretable space and calculate their similarity.
Finally, we compute the EER and AP with respect to the ground-truth scores. We repeat this process for each clustering algorithm and set of parameter values. Figure~\ref{fig:clustering_performance} shows the best performance obtained by each algorithm and the corresponding number of clusters. We select the algorithm that gives us high-performance gain with a lower number of clusters. As performance improvements plateau after DBSCAN, we select it as our clustering algorithm.

\paragraph{Assigning Writing Styles}

\begin{figure}[t]
    \includegraphics[width=0.48\textwidth]{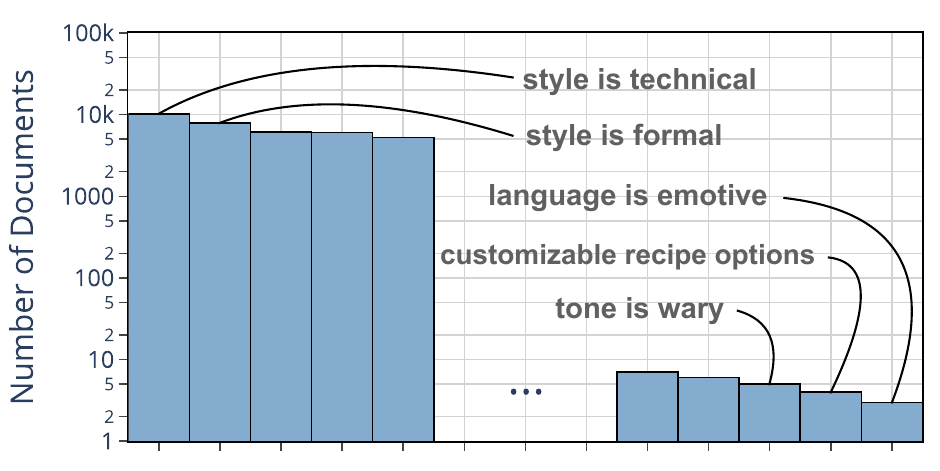}
    \caption{\label{fig:style_corpus} Illustration of style features extracted from the training corpus using our approach. 1470 features were extracted for our dataset.}
\end{figure}

We implement our pipeline for extracting style features from the training documents using the \llama model (Prompt 1 in Figure \ref{fig:prompts}) for style description generation and \gptthree for shortening these descriptions (Prompt 2 in Figure \ref{fig:prompts}). This process resulted in 1,470 style features ranging from generic features (such as ``passive voice''), to corpus-specific ones (such as ``mathematical equations''). 
We include a sample histogram of the extracted style features in Figure~\ref{fig:style_corpus}.

\begin{figure}[ht]
\centering
\begin{minipage}{.45\textwidth}
\begin{lstlisting}[basicstyle=\ttfamily\tiny, frame=single, breaklines=false, caption={Style Generation with \texttt{llama3-8b}}, label={lst:style_features_generation}]
[TASK]
Please list the writing style attributes of the given 
text for each of the morphological, syntactic, semantic, 
and discourse levels.

Each level should start with a paragraph heading then 
a list of short sentences describing the style where each
sentence is in the format of "The author is X." or 
"The author uses X.
[TEXT]
<document>
\end{lstlisting}

\begin{lstlisting}[basicstyle=\ttfamily\tiny, frame=single, breaklines=false, caption={Style Refinement with \texttt{gpt-3.5-turbo}}, label={lst:style_refinement}]
[TASK]: Please rewrite the following list of writing 
style bullet points into a single paragraph.

[TEXT]: <style descriptions>
\end{lstlisting}
        
\end{minipage}
\caption{Prompts used for style description generation and refinement}
\label{fig:prompts}
\end{figure}

\subsection{Baselines}

As explained in Section \ref{sec:approach}, a potential top-down approach to constructing interpretable spaces starts from a pre-defined set of style features and their corresponding documents. We implement this approach and use it as a baseline. We use two existing sets of style features: \xslue \cite{kang:2019} and \lisa \cite{patel-etal-2023-learning}.

\paragraph{\xslue}
\citet{kang:2019} compiled a taxonomy of 18 style features that are mostly studied in literature. A set of texts illustrating each feature is also provided. Due to the varying number of documents available for each feature, we downsampled each set to 100 texts.

\paragraph{\lisa}
\citet{patel-etal-2023-learning} prompted an LLM to distill style features from Reddit posts, resulting in a large collection of style features with corresponding texts. We semi-automatically aggregated similar features and filtered out infrequent ones, resulting in 57 style features with an average of 377 documents per feature.

We compute the overlap between the style features present in each of these corpora and ours. As shown in Figure \ref{fig:style_overlap} in the appendix, the percentage of overlap is low, indicating the need for style feature discovery in the AA model's training corpora instead of relying on pre-defined ones.

To construct the interpretable space for each of these two feature sets, we obtained the AA model's representation for the texts that are available for each style feature and averaged their embeddings to form the corresponding dimension. This results in two baseline interpretable spaces: \xslue with 18 dimensions and \lisa with 57 dimensions. In addition to considering \xslue and \lisa baselines, We also use a random projection baseline by randomly selecting points in the latent space and considering them as another baseline interpretable space.

\section{Evaluation}\label{sec:evaluation}
\begin{table}[t]
	\centering
        \setlength{\tabcolsep}{4pt}
        \scalebox{0.9}{
        \begin{tabular*}{\linewidth}{llrrr}
            \toprule
            \bf Model & \bf Method  & $\Delta$\bf EER & $\Delta$\bf AP & \bf Pearson $r$\\
            \midrule
            \multirow{4}{*}{\bf LUAR} & Random & 0.18 & 0.15 & 0.31\\
                                      & \xslue & 0.17 & 0.27 & 0.39\\
                                      & \lisa  & 0.10 & 0.18 & 0.56\\
                                      & Ours & \bf 0.03 & \bf 0.14 & \bf 0.79\\
            \midrule
            \multirow{4}{*}{\bf FT-LUAR} & Random & 0.20 & 0.19 & 0.38\\
                                          & \xslue & 0.20 & 0.34 & 0.29\\
                                          & \lisa  & 0.14 &  0.24 & 0.23\\
                                          & Ours & \bf 0.04 & \bf 0.16 & \bf 0.79\\
            \bottomrule
        \end{tabular*}
        }
        \caption{Performance degradation of the LUAR and FT-LUAR models when prediction is performed in the \xslue, \lisa, random projection, and our interpretable spaces, measured by Equal Error Rate ($\Delta$EER) and Average Precision ($\Delta$AP). Pearson $r$ represents the agreement between the AA model's predictions in the latent and interpretable spaces. Lower values of $\Delta$EER and $\Delta$AP is better.} 
        \label{table:performance_recovert_results}
\end{table}

The following presents the automatic and manual evaluations we perform to evaluate our approach.

\subsection{Interpretable Space Alignment}
\label{sec:interpretable_space_alignment}

We evaluate the quality of the interpretable space by assessing how well it is aligned with the original latent space.  
We measure the alignment of the two spaces in terms of performance degradation and prediction agreement when each space is used for prediction. Concretely, at test time, we first embed documents into the latent space and project them into the evaluated interpretable space. We then predict AA scores using the cosine similarity of each pair of documents in each space. 
Finally, we compute the corresponding EER and average precision scores and measure the performance degradation. Similar to \citet{simhi:2023}, we also report the correlation in terms of Pearson's $r$ between the predictions made in the latent and the evaluated interpretable space.

Table~\ref{table:performance_recovert_results} presents our evaluation results. A first observation is that our method results in a minimum drop in both evaluation measures compared to the baselines. 
For prediction agreement, the LUAR model with the \xslue and \lisa interpretable spaces outperforms the random baseline, but this trend does not hold for FT-LUAR. 
However, in both LUAR and FT-LUAR, our proposed interpretable space achieves the best agreement, with the highest correlation (Pearson's $r$ of 0.79) to the original latent space predictions. 
These positive results provide empirical support for the relevance of our clustering-based interpretable space with respect to the AA model's predictions.

\begin{figure}[t]
    \centering
    \includegraphics{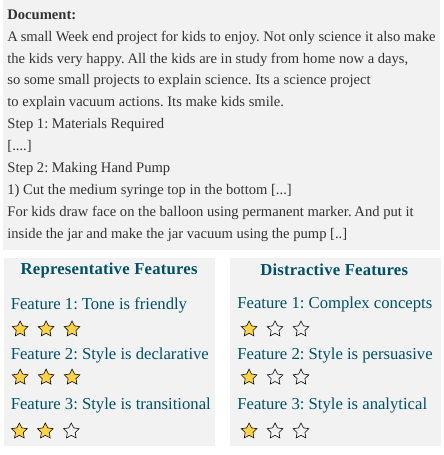}
    \caption{Example document from the HRS dataset with top three descriptive and distractive style features and their scores that are obtained from one of the annotators.}
    \label{fig:exampl-study-1}
\end{figure}

\subsection{Quality of Explanations} 
Our interpretable space can also serve to describe the writing style of unseen documents. We evaluate the quality of the style descriptions for 40 unseen documents randomly sampled from the test split of the HRS dataset. For each document, we extract the top 5 features from its most representative cluster (interpretable dimension) and 5 other style features from the least representative one (distractive features). We then ask human evaluators to rate each of the ten style features on a 3-Point Likert scale. 
They are asked to assign 1 point when the feature does not apply anywhere in the text, 2 points for features that apply somewhere in the text but not frequently, and 3 for features that occur often enough in the text. Ideally, we expect the average score of representative features to be higher than the score of the distractive ones for all cases. Due to a limited budget, we focused our manual evaluation studies on FT-LUAR as our black-box authorship attribution model since it performs better on the HRS dataset. We hired three annotators who are native speakers of English and have a job success rate of more than 90\% on the UpWork platform. Solving each instance was estimated to take around 3 minutes, and we compensated each of our annotators \$35 per hour. We designed the study interface on the Label Studio platform.\footnote{\url{https://labelstud.io/}} 
Figure \ref{fig:exampl-study-1} shows an example instance from the study with the top 3 style features due to space limitations. Note that, in the interface, annotators see all the style features shuffled without knowing their source.

\begin{table}[t!]
    \setlength{\tabcolsep}{8pt}
    \centering
    \begin{tabular}{lrr}
        \toprule
        \bf Features & \bf Average & \bf Median \\
        \midrule
        \bf Representative & \bf 2.41 & \bf 3 \\
        \bf Distractive    & 2.10 &     2\\
        \bottomrule
    \end{tabular}
    \caption{\label{tab:user_study_1_results}
    Average and median rating scores for the representative and distractive features according to the annotators.}
\end{table}

\paragraph{Results} 
We collect 3  ratings from our annotators for each of the 400 evaluated style features. We compute the majority rating for each style feature and exclude the ones with no majority ratings (8\% of representative and 12\% of distractive features). Table \ref{tab:user_study_1_results} presents the average and median scores. We note that representative style features (average 2.41 and median of 3) are scored higher than distractive ones (average 2.10 and median of 2). 

We additionally computed the percentage of documents where the representative features achieved an average rating score higher than the distractive ones. The percentage is 65\% and 72\% when considering the top 3 and 5 evaluated features, respectively. As for the inter-annotator agreement, Krippendorf's $\alpha$ score is 0.33, which indicates fair agreement. Although evaluators found the distractive features sometimes apply, our results indicate the representative style features selected for unseen documents describe the writing style more definitively. In the next section, we show that these style features are also helpful in solving the authorship attribution task. 

\subsection{Usefulness for Explainability}

\paragraph{Study Design} 
\begin{figure*}
	\includegraphics{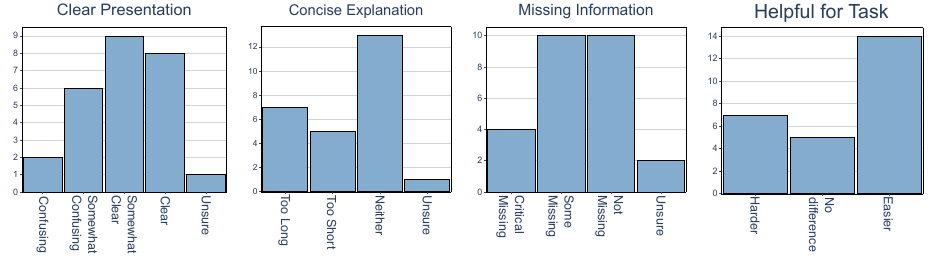}
	\caption{Histogram of annotators' answers for questions regarding the quality aspects of our system's
 explanations.}
	\label{fig:study_2_questions_hist}
\end{figure*}

We conduct another user study to assess whether interpretations derived from our approach can serve as useful explanations for humans to solve the AA task. We hereby present the design of the AA task. We show the annotators one query document along with three candidate documents, one of which is written by the same author as the query. The annotators' task is to identify this candidate document under two conditions:  with and without access to our explanations. 

To generate the explanation for a document, we select the top 10 style features from each of the top three representative interpretable dimensions identified for this document. 
We prompt ChatGPT to rephrase these style features into a single coherent style description. Due to the high annotation effort (estimated 12 min per instance), we restricted our study to a random sample of 40 instances total, 20 for each condition (w/ Expl. and W/o Expl). We hired four annotators on the UpWork platform and split them into two groups; each group solves the same 20 tasks. Additionally, we keep track of the instances where the prediction of the authorship attribution model is correct. Besides solving the AA task, we ask the annotators to answer questions that serve to assess different quality aspects of the provided explanations, such as whether the explanations are clear, compact, complete, and helpful for the task. The questions and the annotation interface can be found in Appendix~\ref{app:user-study-2}.

\begin{table}[t!]
    \small
    \setlength{\tabcolsep}{10pt}
    \centering
    \begin{tabular}{lrr}
        \toprule
        & \bf W/o Explanation & \bf W/ Explanation \\
        \midrule
        \bf Ann 1 & 0.71 & 0.83\\
        \bf Ann 2 & 0.43 & 0.83\\
        \bf Ann 3 & 0.71 & 0.71\\
        \bf Ann 4 & 0.43 & 0.71\\        
        \bottomrule
    \end{tabular}
    \caption{\label{table:user_study_2_results}  Accuracy of each annotator in solving the AA task for instances where the AA model had correct prediction, broken down for cases with and without explanations.}
\end{table}

\paragraph{Effect of Explanations on Accuracy} Table \ref{table:user_study_2_results} presents the accuracy of our annotators in the AA task for cases where the AA model correctly predicted the candidate document. We observe that annotators achieve an accuracy between 0.43 and 0.71 in the setting where no explanations were provided (W/o Explanation).  
When they have access to our explanations (W/ Explanation), the accuracy of most annotators (3 out of 4) significantly increases. 
These results demonstrate the usefulness of our system's explanations for the AA task. 
For instances where the AA model made the wrong prediction (20\% of the total evaluated instances), the annotators also failed to identify the correct candidate in both scenarios (w/ Expl. and w/o Expl.). This indicates that the explanations provided for these instances had no effect or misled the annotators' decision.

\paragraph{Effect of Explanations on Agreement} We collect two annotations per instance, so we measure the inter-annotator agreement using Cohen's Kappa. Agreement across all evaluated instances is low (0.24), hinting at the task's difficulty. Upon further inspection, we observed that agreement was very low (0.03) when no explanations were provided but increased to 0.45 when the annotators had access to the explanations. These results show that our system's 
explanations effectively influenced the annotators' decisions, leading to stronger agreement.

\paragraph{Explanation Quality} 
Figure \ref{fig:study_2_questions_hist} shows the histogram of annotators' answers to our questions regarding the quality of provided explanations. Our system's explanations were evaluated in most cases to be relatively clear, compact, and helpful to the task. However, annotators also considered some information to be missing from the explanations.
\section{Conclusion}
State-of-the-art authorship attribution models learn latent embeddings that capture authors' writing styles. 
Despite their strong performance, these models are unexplainable, which limits their usability. This work presented a novel approach to explaining the latent space of authorship attribution models. Our approach relies on clustering training documents in the latent space, and automatically mapping them to distributions over LLM-generated style attributes. 
In our automatic and manual evaluation, we demonstrated that this method generates plausible style descriptions of unseen documents, which can also be useful in solving the authorship attribution task.
\section*{Limitations}\label{sec:limitation}

In our experiments, we demonstrate the predictive power of our interpretable space, the plausibility of style explanations, and their usefulness for solving the authorship attribution task. However, there exist many other aspects of explanation quality that we did not evaluate \cite{zhou:2021}, like how faithful are these explanations to the model's prediction.

Moreover, as explained, our approach is based on automatically distilling style features using LLMs. This, however, is prone to noise and consistency issues. In Appendix~\ref{app:style_consistency}, we 
investigate the consistency of LLMs in generating writing style descriptions, 
and show that this becomes more reliable when we prompt them multiple times. Due to the computational costs, in our experiments we only perform single-round prompting of LLMs to generate writing style features.  

Finally, in our evaluation we focused on analyzing a single authorship attribution model fine-tuned on our own dataset. This, in a way, is a limitation of the paper; future research should look into analyzing a wider spectrum of authorship attribution models that are trained on different datasets and analyze their different behaviors to gain more reliable and generalizable insights.

\section*{Ethical Statement}
We acknowledge that the authorship attribution task itself raises ethical issues. As mentioned, AA models can be used as tools to reveal the identity of individuals who wrote texts online, leading to privacy concerns. However, our work here aims to explain their decisions. This can enable users to understand these models' behavior and know whether their predictions are baseless.

In our user studies, we made sure to keep the identity of our users private and to compensate them more than the minimum wage in the U.S.

\section{Acknowledgement}
This research is supported in part by the Office of the Director of National Intelligence (ODNI), Intelligence Advanced Research Projects Activity (IARPA), via the HIATUS Program contract \#2022-22072200005. The views and conclusions contained herein are those of the authors and should not be interpreted as necessarily representing the official policies, either expressed or implied, of ODNI, IARPA, or the U.S. Government. The U.S. Government is authorized to reproduce and distribute reprints for governmental purposes notwithstanding any copyright annotation therein.

\bibliography{citations}

\appendix
\begin{figure*}[t]
    \centering
    \begin{minipage}[t]{0.4\textwidth}
        \centering
        \includegraphics[width=\textwidth]{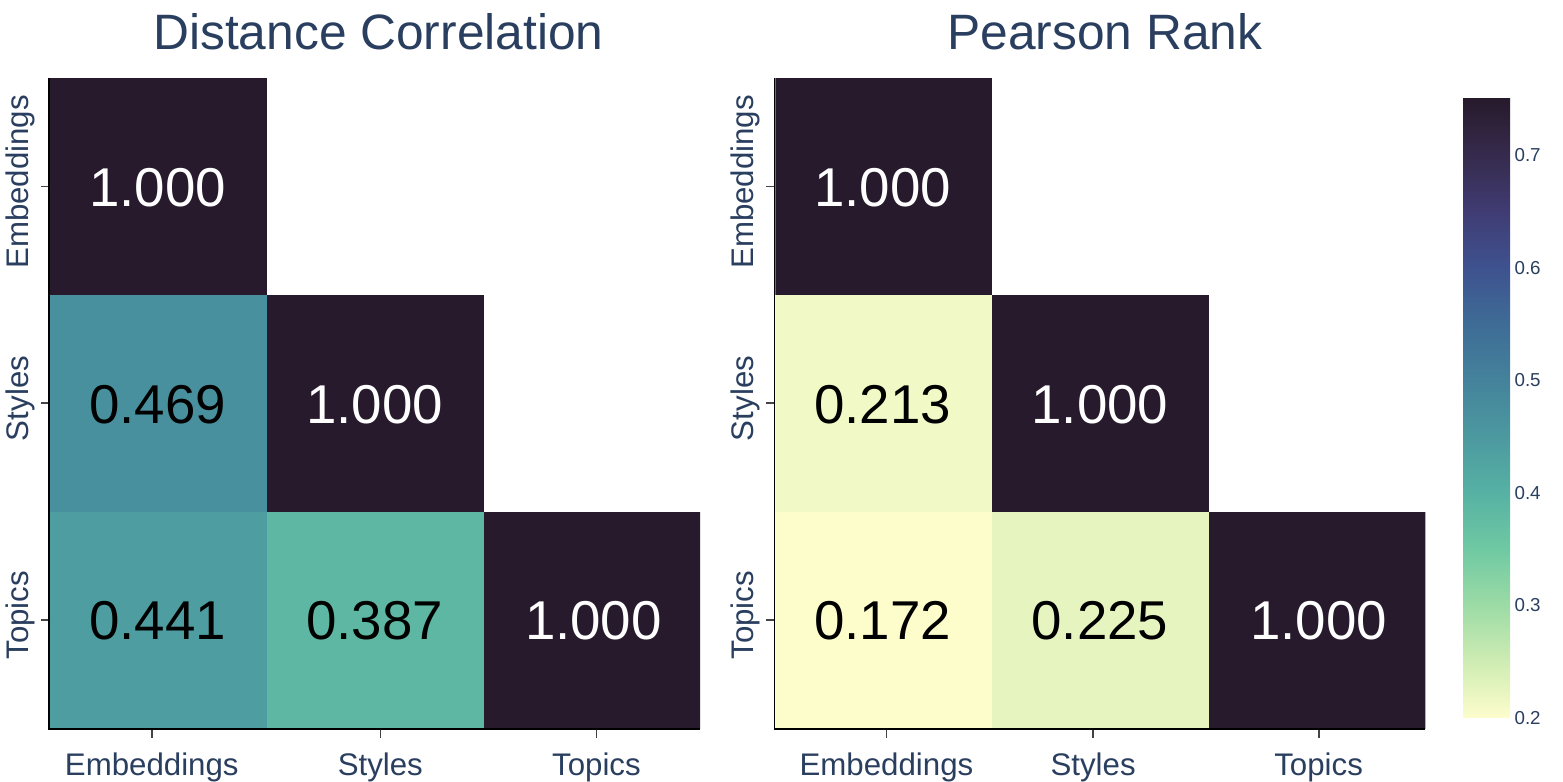} 
        \caption{Correlation statistics. Cosine similarity is used for embeddings and symmetric KL divergence is used for style and topic representations.}
        \label{fig:style_topic_correlation}
    \end{minipage}
    \hfill
    \begin{minipage}[t]{0.3\textwidth}
        \centering
        \includegraphics[width=\textwidth]{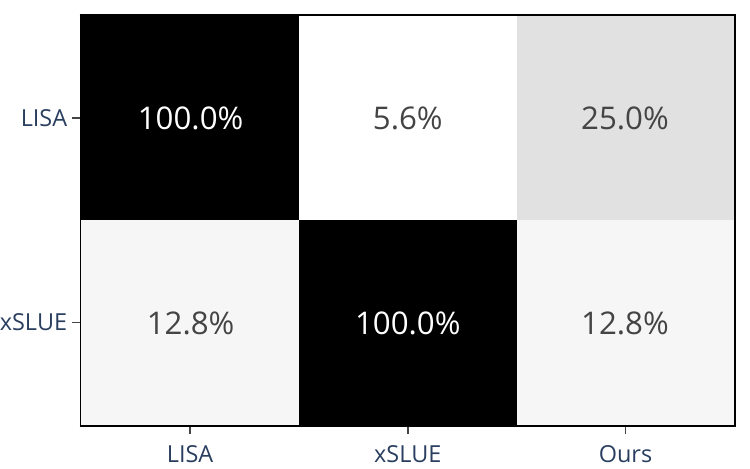}
        \caption{Illustration of style overlap between \lisa, \xslue, and our generated styles.}
        \label{fig:style_overlap}
    \end{minipage}
    \hfill    
    \begin{minipage}[t]{0.23\textwidth}
        \centering
        \includegraphics[width=\textwidth]{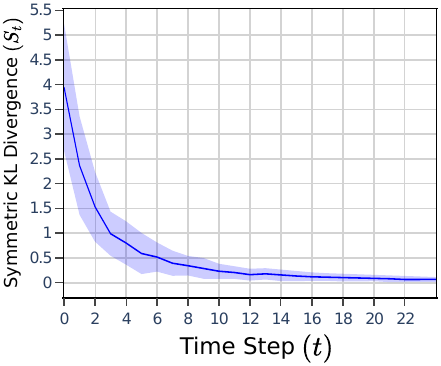} 
        \caption{Evolution of stability $S_t$ across time steps $t$.  Error bands represent standard deviation.}
        \label{fig:style_consistency}
    \end{minipage}
\end{figure*}

\section{Supplementary Analysis}\label{app:analysis}
\subsection{What do Authorship Attribution Models Represent?}\label{app:style_topic_entanglement}

Recent studies observed that authorship attribution models learn representations that capture not only the writing style but also semantics of texts \citep{rivera-soto-etal-2021-learning, wang-etal-2023-authorship}.  Therefore, in a follow-up study, we investigate this phenomena in FT-LUAR by computing correlation scores between the latent embedding on the one hand and style and topic representations on the other hand. Concretely. for topic representations, we use BERTopic \citep{grootendorst2022bertopic} to compute the topic representation of each document and average that to get the final topic representations. We compute the authorship attribution latent representation by averaging the author's document embeddings generated by the FT-LUAR model, and apply the method described in Section~\ref{sec:latent_space_interpretation} to obtain style representations. We then compute pairwise dissimilarities between authors using cosine similarity for latent representations and symmetric KL divergence for style and topic representations. Finally, to obtain correlation scores, we use distance correlation \citep{Sz_kely_2007}, which captures both linear and non-linear relationships, and Pearson correlation, which captures linear relationships.


Results are shown in Figure~\ref{fig:style_topic_correlation}. 
Despite the relatively high correlation between authors' dissimilarity scores in the latent topic space, we observe an even higher correlation between dissimilarities in the latent and style representations. 
This indicates that authors who are similar in the authorship attribution latent space share, to some degree, similar writing styles.

\subsection{Consistency of Style Assignments}\label{app:style_consistency}
To ensure the generated style descriptions are reliable and not spuriously generated, we validate the consistency of style descriptions produced by LLMs across different query instances. 

Specifically, we prompt \llama with documents from the same author multiple times and evaluate the stability of the resulting style distribution. 
At each time step $t \in \mathbb{N}$, we compute the style distribution $v^d_t \in \Delta^k$ by counting the number of times each style occurs and normalizing the occurrences to a probability distribution. Then, we take the unweighted average of all previously generated distributions as $\bar{v}^d_t = \frac{1}{t}\sum_{i=1}^t v^d_i$. 
Denoting stability at step $t$ as $S_t$, we evaluate $S_t$ for $t \leq 25$ as the symmetrized Kullback-Leibler (KL) divergence between the averaged style distributions of step-adjacent instances:
\begin{equation*}
    S_t = \frac{1}{2}\left(D_{\text{KL}}(\bar{v}^d_{t} \parallel \bar{v}^d_{t-1}) + D_{\text{KL}}(\bar{v}^d_{t-1} \parallel \bar{v}^d_{t}) \right).
\end{equation*}

To mitigate potential consistency illusions from using identical prompt instructions, we manually generate 20 paraphrased variants of the original instruction and sample an instruction uniformly at random for each query. Given the known variability of LLMs when processing semantically identical inputs \cite{jiang-etal-2020-know, elazar-etal-2021-measuring, zhou2022promptconsistencyzeroshottask}, observing convergence in style assignments across these varied prompts suggests that LLMs maintain consistency in generating style descriptions. Detailed prompt examples are provided in Appendix \ref{app:style_consistency_prompts}.

Results are shown in Figure~\ref{fig:style_consistency}. Observe that stability converges to the minimum value after 20 rounds, demonstrating that LLM-generated styles are consistent across multiple prompting instances. These results suggest that author-level style assignments are both representative and consistent.

\subsection{Style Overlap}\label{app:style_overlap}
Figure~\ref{fig:style_overlap} shows the heatmap of style overlap between various style corpora. The style overlap is determined by calculating the Mutual Implication Score for all pairs of style features and considering two styles from different corpora identical when the similarity score exceeds a threshold of 0.8. We notice the low percentage overlap between the other baseline corpora and ours, which indicates the need for style feature discovery in the training corpora of the investigated AA model instead of relying on a pre-defined set of style features.

\section{Prompt Instructions}
\subsection{Style Feature Generation}\label{app:style_filtering}
Here, we include the prompt instructions used to generate style descriptions and refine the generated descriptions.

\subsection{Style Consistency}\label{app:style_consistency_prompts}
Here, we include examples of paraphrased style generation prompts utilized in assessing the consistency of generated styles.

\begin{figure}[ht]
\centering
    \begin{minipage}{.45\textwidth}
        \begin{lstlisting}[basicstyle=\ttfamily\tiny, frame=single, breaklines=false, caption={Examples of Paraphrased Variants}, label={lst:paraphrased_style_generation}]
[TASK]
Generate a list detailing the writing style of the given 
text at the following levels:
 - Morphological level
 - Syntactic level
 - Semantic level
 - Discourse level
  
[RULES]
 - Begin each level with a heading, followed by a list 
   of brief sentences describing the style.

[TEXT]: <document>
        \end{lstlisting}
        \begin{lstlisting}[basicstyle=\ttfamily\tiny, frame=single, breaklines=false]
[TASK]
Assess the writing style of the given text for each of 
the following levels:
 - Morphological level
 - Syntactic level
 - Semantic level
 - Discourse level
  
[RULES]
 - Each section should start with a heading, followed
   by a list of brief sentences describing the style.

[TEXT]: <document>
        \end{lstlisting}
        
    \end{minipage}
\end{figure}

\section{Experiment on Usefulness of Explanations}\label{app:user-study-2}

Besides solving the task of authorship attribution, the annotators had to answer a set of questions for each instance where explanations were provided. These questions meant to assess various quality aspect of our explanations. Here is the list of questions:
\begin{enumerate}
    \item Is the explanation presented in a clear manner that is easy to follow?
        \subitem A. Confusing or difficult to understand
        \subitem B. Somewhat confusing or difficult to understand
        \subitem C. Relatively clear and easy to understand
        \subitem D. Very clear and easy to understand
        \subitem E. I am not sure

    \item Is the explanation simple and compact?
        \subitem A. Too long
        \subitem B. Too short
        \subitem C. Not too long or too short
        \subitem D. I am not sure

    \item Is there any missing information that the explanation fails to mention?
        \subitem A. Critical information is missing
        \subitem B. Some information is missing, but not critical
        \subitem C. No information is missing
        \subitem D. I am not sure
        
    \item Is the explanation helpful to the task you are trying to accomplish?
        \subitem A. The explanation is distracting
        \subitem B. The explanation makes little or no difference to my task
        \subitem C. The explanation is helpful
        \subitem D. I am not sure
\end{enumerate}

\end{document}